\documentclass[11pt]{article}

\usepackage{acl}

\usepackage{times}
\usepackage{latexsym}
\usepackage{bbm}

\usepackage[T1]{fontenc}

\usepackage[utf8]{inputenc}

\usepackage{microtype}

\usepackage{inconsolata}

\usepackage{graphicx}
\usepackage{amsmath}
\usepackage{cleveref}
\usepackage{booktabs}
\title{When to Think Deeply: Inhibitory Deliberation for LLM Reasoning}

\author{\bf{Zhixuan He \quad Yue Feng}\\
hezhixuan1997@gmail.com\\
y.feng.6@bham.ac.uk\\
University of Birmingham, United Kingdom
}

\begin{document}
\maketitle
\begin{abstract}
Reasoning Large Language Models can improve problem-solving performance through deliberative inference, but invoking slow reasoning for every input is computationally expensive and often unnecessary. We propose IDPR, a framework for response-conditioned inhibitory deliberation. IDPR first generates a concise intuitive answer and then uses an inhibition controller to decide whether that specific response should be released or suppressed in favor of slow reasoning. Unlike input-only routers, the inhibition controller conditions on the fast answer and fast-side evidence, including confidence, logit margin, parseability, and generation cost. We train the controller from paired fast-slow outcomes and select the inhibition threshold on a held-out validation set under an accuracy-first slow-call budget. On a held-out 5,000-example mathematical reasoning test set, IDPR invokes slow reasoning on only 8.20\% of examples and improves accuracy from 47.90\% to 48.92\%. Under the same slow-call budget, random routing decreases accuracy to 46.76\%, while the strongest confidence-based baseline reaches 48.22\%. IDPR also achieves the highest corrective precision, showing that response-conditioned inhibition better identifies fast answers that benefit from slow reasoning.
\end{abstract}

\section{Introduction}
Recent progress in foundation agents suggests that future AI systems should not be viewed merely as monolithic language models, but as modular cognitive systems that combine reasoning, memory, perception, control, and action. A prominent direction is to design agents with brain-inspired architectures that draw on cognitive science and neuroscience, where different modules support different functional roles such as reasoning, world modeling, reward processing, and control\cite{liu2025foundationagents}. At the same time, reasoning-specialized language models have made deliberative inference a core capability of the modern LLM system\cite{deepseekai2025deepseekr1}, making slow reasoning increasingly powerful but also more computationally expensive. These developments raise an important question: \textbf{if slow reasoning is powerful but costly, when should an intelligent system invoke it?}

A natural answer is not deliberate all the time. In human cognition, fast intuitive responses often arise before slower reasoning. Executive control can suppress such prepotent responses when conflict, error risk, or task demands indicate that additional control is needed. However, cognitive control is itself costly; theories of the expected value of control suggest that control should be recruited selectively when its expected benefit justifies the effort\cite{botvinick2001conflict,shenhav2013expected}. This perspective is especially relevant for LLM reasoning. A fast model may often produce a short correct answer at low cost, while a reasoning model may spend hundreds or thousands of tokens on slow reasoning. \textbf{The key challenge is therefore not simply how to do slow reasoning, but how to decide whether a particular fast response should be trusted or inhibited.}

We propose \textbf{IDPR}, a framework of \textbf{Inhibitory Deliberative Problem Reasoning}. The central distinction of IDPR is that routing is response-conditioned. IDPR first elicits a concise intuitive response from a fast policy and treats it as prepotent candidate answer. An inhibition controller then observes the question, the fast answer, and fast-side evidence such as confidence, logit margin, parseability, and generation cost. The controller estimates whether invoking a slow reasoning policy is expected to provide sufficient benefit over the fast response. If the expected benefit is high, IDPR suppresses the fast answer and invokes the slow policy; otherwise, it releases the fast answer directly.

In this work, we evaluate IDPR in an accuracy-first selective deliberation setting, where the goal is to improve correctness under a bounded slow-call budget while transparently reporting inference cost. We report accuracy, actual slow-call rate, generated-token cost, and cost-aware utility. To isolate the quality of the routing decision, we compare against same budget baselines that invoke slow reasoning on the same number of examples as IDPR, including random slow-call selection and confidence-based routing using fast-model log probability and logit margin. 

On a held-out mathematical reasoning test set, IDPR invokes slow reasoning on only 8.20\% of examples and improves accuracy from 47.90\% to 48.92\%. Under the same slow-call budget, random routing decreases accuracy to 46.76\%, while the strongest confidence-based baseline reaches only 48.22\%. IDPR also achieves the highest corrective precision: 27.07\%, indicating that the learned inhibition controller better identifies fast responses that benefit from slow reasoning.

Our contributions are as follows:
\begin{itemize}
    \item We introduce IDPR, an inhibition framework that decides whether a fast answer should be released or suppressed in favor of slow reasoning.
    \item We propose an inhibition controller that estimates slow-over-fast quality gain and slow reasoning cost from question, fast answer, and fast-side evidence.
    \item We evaluate IDPR under an accuracy-first selective slow reasoning protocol with explicit cost reporting, showing that it improves held-out reasoning accuracy and outperforms random and confidence-based routing baselines under the same budget.
\end{itemize}

\section{Related Work}\label{sec:related}
\paragraph{Efficient LLM routing and cascades.}
Recent work has explored cost-aware routing and cascade systems for large language models, where the goal is to trade off response quality against inference cost. FrugalGPT studies LLM cascades that adaptively select among different LLM calls to reduce cost while preserving or improving accuracy~\citep{chen2023frugalgpt}. Hybrid LLM routes queries to small or large models based on predicted query difficulty and a desired quality level~\citep{ding2024hybrid}, while RouteLLM learns routers from preference data to dynamically select between weaker and stronger models~\citep{ong2025routellm}. These systems are effective for model selection and cost-quality trade-offs, but they typically make routing decisions before conditioning on an actual candidate response. This pre-response routing formulation leaves out a crucial signal for reasoning tasks: the fast answer itself. In mathematical reasoning, the content, format, confidence, and uncertainty profile of a fast answer can reveal whether it is likely to be correct, malformed, or in need of further deliberation. IDPR therefore adopts a formulation: it first generates a fast answer and then decides whether that specific response should be released or inhibited. Thus, the controller does not merely ask whether the input question appears difficult; it asks whether the particular fast response is worth suppressing in favor of slow deliberation.
\paragraph{Reasoning models and test-time computation.} Recent work has shifted from prompts engineering to building models explicitly optimized for deliberative reasoning. Deepseek-R1 shows that large scale reinforcement learning, cold-start data, multi-stage training, and distillation can elicit strong reasoning behaviors and produce distilled reasoning models at multiple scales\cite{deepseekai2025deepseekr1}. Kimi k1.5 further demonstrates that large-scale reinforcement learning and long-context scaling can improve reasoning performance across multi-modal and text-only tasks\cite{kimi2025k15}. Open-Reasoner-Zero and Skywork-OR1 explore scalable open-source reinforcement recipes for reasoning models, showing that rule-based rewards and long-form reasoning optimization can substantially improve mathematical and coding performance\cite{hu2025openreasonerzero,he2025skyworkor1}. Other work studies data-efficient or curated post-training for reasoning, such as LIMO and Phi-4-reasoning, which show that carefully selected demonstrations or teachable prompts can elicit strong reasoning behavior from capable base models\cite{ye2025limo,abdin2025phi4reasoning}. In parallel, test-time compute scaling studies show that inference-time computation can be controlled or expanded through mechanisms such as budget forcing, repeated sampling, or compute-optimal allocation~\citep{muennighoff2025s1,snell2025scalingtesttime,brown2025largelanguagemonkeys}. IDPR is complementary to these efforts: rather than proposing a new reasoning model or a new way to spend more test-time compute, it learns when a costly deliberative reasoner should be invoked after observing the actual fast response.
\paragraph{Response-conditioned inhibition and cognitive control.}
IDPR is motivated by cognitive accounts of fast intuitive responses and costly deliberative control. Dual-process theories distinguish fast, automatic responses from more controlled reasoning processes~\citep{evans2013dualprocess,kahneman2011thinking}. In cognitive control, an initially activated or prepotent response can be withheld or overridden when task demands indicate that the response may be inappropriate. This idea is closely related to response inhibition, where control mechanisms stop or suppress a dominant action tendency~\citep{logan1984executive,aron2014inhibition}. Conflict monitoring theories further suggest that conflict or expected error risk can signal the need for additional control~\citep{miller2001integrative,botvinick2001conflict}. Importantly, such control is not free: people tend to avoid cognitively demanding actions~\citep{kool2010decision}, cognitive effort carries subjective cost~\citep{westbrook2013subjective}, and the expected value of control framework formalizes control allocation as a trade-off between expected benefit and effort cost~\citep{shenhav2013expected}. IDPR operationalizes this view in LLM inference. The fast answer is treated as a prepotent response that has already been formed, and the inhibition controller decides whether to release it or suppress it in favor of slow reasoning. Unlike simple confidence filtering, the decision is conditioned on the actual fast response and estimates both slow-over-fast quality gain and deliberation cost.

\section{Methodology}\label{sec:method}

\subsection{Problem Setup}
Given an input reasoning problem $x$, our goal is to produce a final answer $y$ under a utility-aware computation budget. We consider two reasoning policies with complementary properties. The fast policy $\pi_{f}$ produces a concise intuitive response:
\begin{equation}
    y_f \sim \pi_{f}(\cdot \mid x),
\end{equation}
while the slow policy $\pi_{s}$ performs a deeper reasoning process and produces:
\begin{equation}
    y_s \sim \pi_s(\cdot | x).
\end{equation}
and the final output is 
\begin{equation}
    y=
\begin{cases}
    y_f, &r=\mathrm{fast},\\
    y_s, &r=\mathrm{slow}.
\end{cases}
\end{equation}
Unlike standard accuracy-oriented reasoning systems, our objective is not simply to maximize the probability of correctness by always using the most expensive reasoning path. Instead, we optimize a utility that trades off answer quality and inference cost. For an output $y$, we define 
\begin{equation}
    U(y,x)=\alpha \cdot A(y,x)-\lambda_{\mathrm{tok}}\cdot C(y)-\rho F(y),
\end{equation}
where $A(y,x)$ measures answer correctness, $C(y)$ denotes the generated token or inference cost, and $F(y)$ penalizes malformed or non-parseable outputs. The coefficients $\alpha$, $\lambda_{\mathrm{tok}}$, and $\rho$ control the relative importance of correctness, computation, and answer format.

In contrast to pre-routing approaches that select a reasoning path solely from the input $x$, our setting conditions the decision on the actual fast response $y_f$ and its associated fast-side evidence. This formulation allows the system to suppress a fast intuitive response when additional deliberation is predicted to provide sufficient corrective benefit, while releasing the fast response when it appears reliable or when the expected gain from slow reasoning does not justify its cost.
\subsection{Overall Framework}
\begin{figure*}[t]
    \centering
    \includegraphics[width=\textwidth]{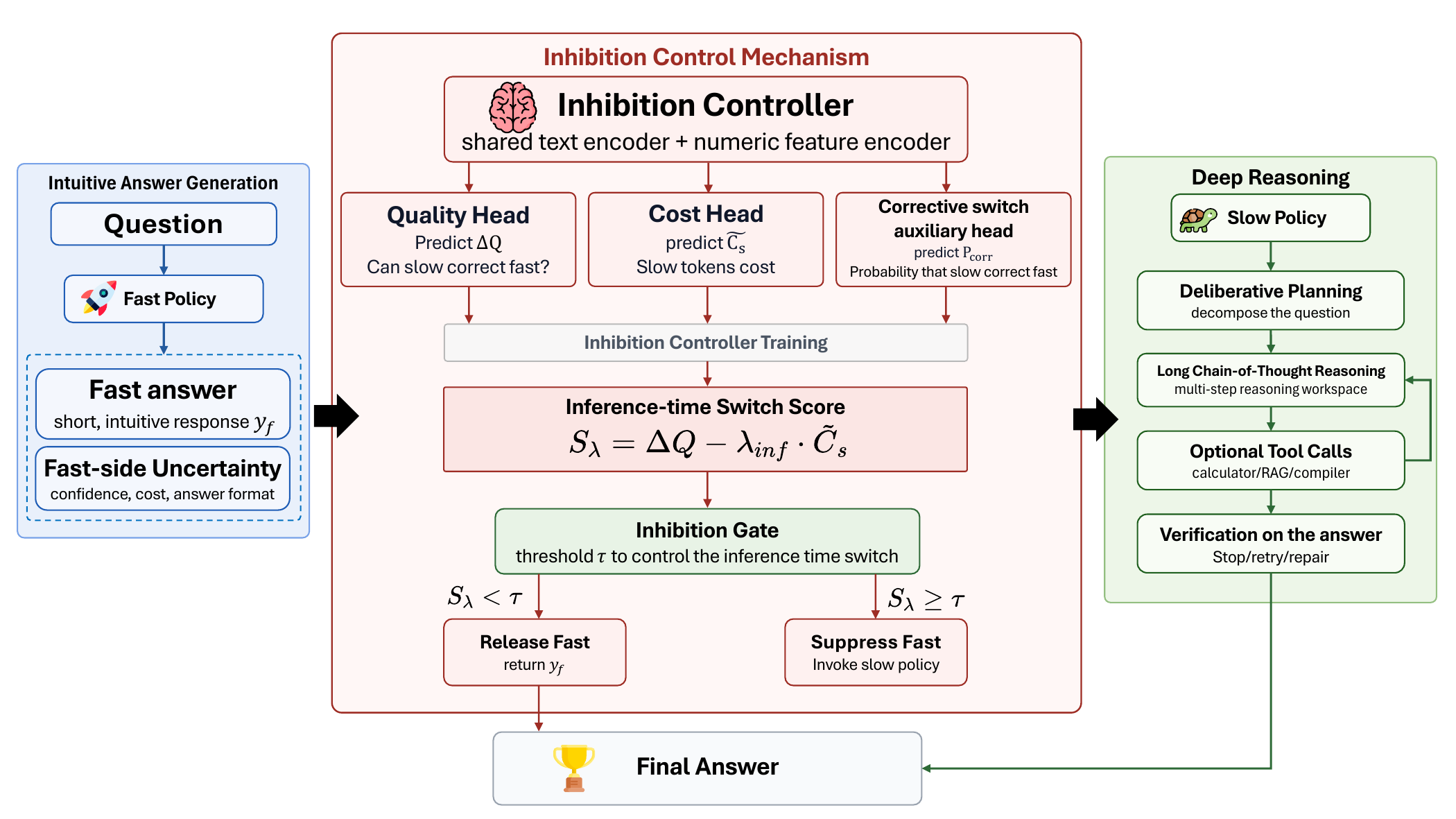}
    \caption{Overview of IDPR. The system first generates an intuitive fast response and then applies an inhibition controller conditioned on the question, the fast answer, and fast-side evidence. If the switch score satisfies $S_\lambda \ge \tau$, the fast response is suppressed and the slow reasoning branch is invoked; otherwise, the fast response is released directly.}
    \label{fig:overview}
\end{figure*}
Figure~\ref{fig:overview} provides an overview of the proposed IDPR framework. IDPR follows an inhibitory reasoning paradigm. Given a question $x$, the system first generates a concise intuitive answer $y_f$ using the fast policy $\pi_f$. This fast response is treated as a prepotent candidate answer: by default, it can be released directly as the final output, but it may also be suppressed when the controller predicts that a deeper reasoning is beneficial.
The routing decision is made by a response-conditioned inhibition controller conditioned on the actual fast response. Specifically, after obtaining $y_f$, we extract a set of fast-side evidence features $c_f$, including confidence, logit margin, generation length, answer form, parseability, latency, and repetition statistics. The controller then receives:
\begin{equation}
    e_f=(x,y_f,c_f),
\end{equation}
The controller estimates the expected quality gain of slow reasoning and its cost, and computes a switch score:
\begin{equation}
    S_{\lambda}=\widehat{\Delta Q}-\lambda_{\mathrm{inf}} \cdot \widehat{C_s},
\end{equation}
where $\lambda_{\mathrm{inf}}$ controls the inference-time sensitivity to computational cost. A larger $S_\lambda$ indicates that slow reasoning is expected to provide a larger net benefit. The inhibition gate is therefore defined as:
\begin{equation}
    r=
\begin{cases}
    \mathrm{slow}, & S_\lambda \ge \tau,\\
    \mathrm{fast}, & S_\lambda < \tau,
\end{cases}
\end{equation}
where $\tau$ is a calibrated threshold. If $S_\lambda < \tau$, the fast response is released as the final answer. If $S_\lambda \ge \tau$, the controller suppresses the fast response and invokes the slow policy $\pi_s$ to produce a deeper reasoning answer.
As discussed in Section~\ref{sec:related}, prior LLM routing and cascade methods typically decide which model or route to use before conditioning on an actual candidate response. IDPR instead makes an inhibition decision after the fast answer has been formed.
\subsection{Intuitive Response Generation}
IDPR begins by eliciting a concise intuitive response from the fast policy. The fast branch is trained or prompted to output a short, parseable final answer rather than a full deep reasoning trace. This makes the fast response inexpensive and allows it to serve as the default candidate answer before any slow reasoning is invoked. The fast response also provides evidence for deciding whether it should be released or inhibited. We extract fast-side evidence:
\begin{equation}
    c_f=\phi(x,y_f,\pi_f),
\end{equation}
which summarizes the reliability, uncertainty, and computational properties of the fast response. In practice, $c_f$ includes four categories of features: confidence features such as average log-probability and logit margin; parseability features such as answer form, candidate count, format status; generation-cost features such as fast token count, latency, and throughput; and degeneracy features such as repetition and truncation.
All fast-side evidence is available at inference time and does not require access to the gold answer. This evidence allows the controller to evaluate the actual fast response, rather than routing solely from the input question. We provide the full feature list and extraction details in

\subsection{Inhibition Controller: Quality and Cost Trade Off}\label{sec:controller}
The inhibition controller decides whether the fast response should be released or suppressed. Given the controller input $e_f=(x,y_f,c_f)$, it combines textual and numerical evidence. A shared text encoder processes the question and fast answer, while a numerical feature encoder processes the fast-side evidence:
\begin{equation}
    h=\mathrm{MLP}(\mathrm{Enc}_{\mathrm{text}}(x,y_f);\mathrm{Enc}_{\mathrm{num}}(c_f)).
\end{equation}
The representation $h$ is shared by three prediction heads: (i)the expected quality gain $\widehat{\Delta Q}$ of invoking slow reasoning, (ii)the slow reasoning cost $\widehat C_s$, and (iii) an auxiliary corrective probability $\hat p_{\mathrm{corr}}$ indicating whether slow reasoning is likely to correct an incorrect fast response.
The quality gain target is defined as
\begin{equation}
    \Delta Q=\alpha(A_s-A_f)-\rho(F_s-F_f),
    \label{eq:quality_gain}
\end{equation}
where $A_f$ and $A_s$ denote the correctness of fast and slow responses, and $F_f$ and $F_s$ denote their format penalties. The cost head is trained on a normalized log-cost target to handle the long-tailed distribution of slow reasoning traces. The auxiliary corrective head provides additional supervision for identifying cases where the fast response should be inhibited.
The switch score is a learned surrogate for the utility difference between invoking the slow branch and releasing the fast response:
\begin{equation}
    S_\lambda=\widehat{\Delta Q}-\lambda_{\mathrm{inf}}\widehat C_s,
\label{eq:switch_score}
\end{equation}
where $\lambda_{\mathrm{inf}}$ controls deployment-time cost sensitivity. The inhibition gate is 
\begin{equation}
    r =
\begin{cases}
    \mathrm{slow}, & S_\lambda \ge \tau,\\
    \mathrm{fast}, & S_\lambda < \tau.
\end{cases}
\label{eq:gate}
\end{equation}
Thus, high switch scores suppress the fast response and invoke slow reasoning, while low scores release the fast answer directly.

\subsection{Deep Reasoning}
When the inhibition gate suppresses the fast response, IDPR invokes the slow policy $\pi_s$ to perform the slow reasoning. The slow branch is not used for every input; it is conditionally activated only when the inhibition controller predicts that slow reasoning is likely to provide sufficient benefit over the fast response. Formally, when $\S_\lambda \ge \tau$, the system generates
\begin{equation}
    y_s \sim \pi_s(\cdot | x),
\end{equation}
and returns the slow response as the final answer.
The slow branch provides a slow reasoning pathway that can decompose the problem, perform multi-step reasoning, and verify the final answer before returning it. In our implementation, the slow policy is prompted to produce a detailed solution followed by a parseable final answer. A lightweight repair step can be applied when the final answer is missing or malformed. This design keeps slow reasoning as a conditional resource: IDPR pays the cost of deep reasoning only when the fast response is predicted to be unreliable or insufficient.

\subsection{Controller Training and Calibration}
The inhibition controller is trained from paired fast-slow outcomes. For each problem, we evaluate the fast response $y_f$ and slow response $y_s$ against the gold answer and use the resulting correctness, format penalty, and slow cost to supervise the three controller heads defined in \cref{sec:controller}. We train the inhibition controller with 
\begin{equation}
    \mathcal L_{\mathrm{ctrl}}=\mathcal L_Q+\beta_C\mathcal L_C+\beta_R\mathcal L_R,
\end{equation}
where $\mathcal L_Q$ and $\mathcal L_C$ are regression losses for quality gain and slow cost, $\mathcal L_R$ is the binary classification loss for the corrective label, and $\beta_C,\beta_R$ control the relative weights of the cost and corrective objectives.
At deployment, the continuous score $S_\lambda$ must be converted into a binary routing decision. We therefore select the threshold $\tau$ on a held-out validation set according to the desired deployment objective. In our setting, we choose the threshold that maximizes validation accuracy under a bounded slow-call budget. The selected threshold is then fixed for test evaluation. This calibration step keeps the actual slow-call rate and inference cost transparent while allowing the system to adapt the inhibition threshold to the deployment objective.

\section{Experiments}\label{sec:experiments}
We evaluate IDPR on held-out mathematical reasoning tasks. Our experiments are designed to answer three questions: (i)whether the inhibition control can improve accuracy under a bounded slow-call budget, (ii)whether the learned inhibition controller selects better slow reasoning targets than random or confidence-based routing method, and (iii)how the additional slow reasoning cost compares with the accuracy gain.
\subsection{Experimental Setup}
\paragraph{Datasets.} We evaluate IDPR on mathematical reasoning problems drawn from two sources. The first source is GSM8K, a benchmark of grade-school math word problems designed to evaluate multi-step arithmetic reasoning\cite{cobbe2021gsm8k}. The second source consists of harder mathematical reasoning problems from the math subsets of Mixture-of-Thoughts(MoT) dataset\cite{openr1}. We construct a router pool containing paired fast and slow outputs for each problem. For final evaluation, we exclude examples used for inhibition controller training and construct a held-out validation set fo 2,000 examples and a disjoint held-out test set of 5,000 examples, balanced across the two source groups, while all main results are reported on the held-out test set.
\paragraph{Fast and slow policies.}The fast policy is an answer-only reasoning model based on Qwen2.5-Math-7B, trained or prompted to produce a concise parseable final answer. Qwen2.5-Math is a math specialized model family with 7B instruction-tuned variants and strong mathematical reasoning capabilities\cite{yang2024qwen25math}. The slow policy is OpenR1-Distill-7B, a reasoning model that replicates the reasoning behavior of Deepseek-R1-Distill-Qwen-7B in an open and reproducible form\cite{openr1distill7b}. We use the slow policy only when the inhibition controller suppresses the fast answer. This choice reflects our goal of studying when to invoke a costly reasoning model rather than training a new slow reasoning model.
\paragraph{Router pool construction.}To train the inhibition controller, we construct paired fast-slow supervision by running both policies on the same question. For each problem $x$, the fast policy produces a concise answer $y_f$, while the slow policy produces a deliberative answer $y_s$. We evaluate both outputs against the gold answer and record correctness, format validity, and token generation cost. These paired outputs define the quality gain of invoking slow reasoning, the slow reasoning cost, and whether slow reasoning corrects an erroneous fast response. Thus, the routing labels are derived automatically from paired fast-slow outputs rather than manually annotated.
\paragraph{Baselines.}Our primary comparisons are same-budget routing baselines. Always-Fast returns the fast answer for all examples. Random Same-k invokes the slow policy on the same number of test examples as IDPR but chooses them uniformly at random. Low Avg LogProb Same-k and Low Gap Same-k also use the same number of slow calls, but choose examples with the lowest fast model confidence, measured by average token log-probability and average logit margin. These baselines isolate the quality of the routing decision under the same slow-call budget. 
\paragraph{Metrics.} Accuracy is the primary metric. Since IDPR explicitly trades additional slow reasoning for improved correctness, we also report slow-call rate and average generated tokens. We report the cost-aware utility defined in \cref{sec:method} as a diagnostic measure. For switched examples, we report corrective precision, the fraction of routed examples for which the fast answer is incorrect and the slow answer is correct.

\subsection{Main Results}
\begin{table*}[t]
\centering
\small
\begin{tabular}{lrrrrrr}
\toprule
Method & Slow Rate & Accuracy & $\Delta$ Acc. & Utility & Avg. Tokens & Corr. Prec. \\
\midrule
Always-Fast & 0.00\% & 47.90 & -- & 0.4789 & 5.26 & -- \\
Random Same-k & 8.20\% & 46.76 & -1.14 & 0.4517 & 517.91 & 13.41 \\
Low Avg LogProb Same-k & 8.20\% & 48.22 & +0.32 & 0.4646 & 582.68 & 21.22 \\
Low Logit Gap Same-k & 8.20\% & 47.84 & -0.06 & 0.4571 & 714.73 & 11.46 \\
\textbf{IDPR} & \textbf{8.20\%} & \textbf{48.92} & \textbf{+1.02} & \textbf{0.4731} & \textbf{519.42} & \textbf{27.07} \\
\bottomrule
\end{tabular}
\caption{
Main held-out test results. Same-k baselines invoke slow reasoning on the same number of examples as IDPR. IDPR achieves the highest accuracy and corrective precision among routing methods, showing that the learned inhibition controller selects more useful deliberation targets than random or confidence-based routing.
}
\label{tab:main_results}
\end{table*}
Table~\ref{tab:main_results} shows the main held-out test results. IDPR invokes slow reasoning for 8.20\% of examples and improves accuracy from 47.90\% to 48.92\%. Under the same slow-call budget, random routing substantially decreases accuracy, while confidence-based routing provides smaller gains. IDPR also achieves the highest corrective precision, indicating that the learned inhibition controller better identifies fast responses that benefit from slow reasoning.

The comparison with Random Same-k shows that simply spending the same amount of slow reasoning computation is not sufficient: random routing decreases accuracy by 1.14 points relative to Always-Fast. Confidence-based routing is stronger, especially Low Avg LogProb Same-k, but it still underperforms IDPR by 0.70 points and has lower corrective precision. This suggests that the inhibition controller learns more than a generic uncertainty heuristic. It estimates whether slow reasoning is likely to provide a corrective benefit for the particular fast response.

As expected for an accuracy-first strategy, IDPR incurs additional token cost and does not maximize the cost-aware utility relative to Always-Fast. This trade-off is consistent with the cognitive motivation of IDPR. Deliberative control in human cognition is also costly and is recruited selectively when the expected benefit justifies the effort\cite{kool2010decision,shenhav2013expected}. Similarly, switching from fast to slow reasoning in LLMs incurs additional token cost, but can be justified in an accuracy-first setting when it improves correctness on difficult examples. We therefore report slow-call rate and average generated tokens explicitly rather than treating slow reasoning as free. What's more, under the same slow-call budget, IDPR obtains substantially higher utility than all routing baselines, indicating that its selected slow calls are more useful than those chosen by random or confidence-based policies.

\section{Discussion}
We further discuss where IDPR gains come from and how different design choices affect performance. We first break down the results by source group to understand whether selective slow reasoning is more useful on harder problems. We then compare inhibition controller and fast policy variants, including sampled fast controller augmentation and GRPO fast policy refinement.
\subsection{Source wise Analysis}
\begin{table*}[t]
\centering
\small
\begin{tabular}{llrrrr}
\toprule
Source & Method & Slow Rate & Accuracy & $\Delta$ Acc. & Avg. Tokens \\
\midrule
GSM & Always-Fast & 0.00\% & 60.20 & -- & 5.31 \\
GSM & Low Avg LogProb & 8.52\% & \textbf{61.04} & +0.84 & 481.30 \\
GSM & IDPR & 6.68\% & 60.96 & +0.76 & \textbf{325.98} \\
\midrule
MoT & Always-Fast & 0.00\% & 35.60 & -- & 5.21 \\
MoT & Low Avg LogProb & 7.88\% & 35.40 & -0.20 & 684.07 \\
MoT & IDPR & 9.72\% & \textbf{36.88} & \textbf{+1.28} & 712.86 \\
\bottomrule
\end{tabular}
\caption{
Source-wise results. IDPR improves accuracy on both source groups, with a larger gain on the harder MoT subset. Low Avg LogProb is the strongest confidence baseline on GSM, but fails to improve accuracy on MoT.
}
\label{tab:source_wise}
\end{table*}
Table~\ref{tab:source_wise} breaks down performance by source group. IDPR improves accuracy on both GSM8K and MoT, but the gain is larger on the harder MoT subset. On GSM8K, IDPR improves accuracy from 60.20\% to 60.96\%, while using fewer tokens and achieving higher corrective precision than the strongest confidence baseline. On MoT, the Always-Fast baseline accuracy drops to 35.60\%, IDPR improves accuracy to 36.88\%. In contrast, Low Avg LogProb and other baselines do not improve over Always-Fast on MoT.

IDPR also allocates more slow calls to the harder subset: its slow-call rate is 6.68\% on GSM8K and 9.72\% on MoT. This suggests that the inhibition controller does not distribute slow reasoning uniformly, but instead directs more slow reasoning to examples where fast responses are less reliable. The larger gain on MoT supports the intended role of IDPR as a selective slow reasoning mechanism for difficult reasoning cases.
\subsection{Controller and Fast Policy Variants}
\begin{table*}[t]
\centering
\small
\begin{tabular}{llrrrrr}
\toprule
Fast Policy & Controller & Slow Rate & Accuracy & $\Delta$ Acc. & Avg. Tokens & Corr. Prec. \\
\midrule
Answer-only Fast & Deterministic-scan & \textbf{8.20\%} & \textbf{48.92} & \textbf{+1.02} & 519.42 & \textbf{27.07} \\
Answer-only Fast & Sampled-fast & 7.78\% & 48.78 & +0.88 & \textbf{471.38} & 25.71 \\
GRPO-refined Fast & Deterministic-scan & 8.96\% & 48.74 & +1.04 & 554.97 & 26.12 \\
GRPO-refined Fast & Sampled-fast & 7.78\% & 48.56 & +0.86 & 472.59 & 25.71 \\
\bottomrule
\end{tabular}
\caption{
Controller and fast-policy variants on the held-out test set. 
``Answer-only Fast'' denotes the main concise-answer fast policy. 
``GRPO-refined Fast'' denotes the fast checkpoint obtained after GRPO refinement. 
``Deterministic-scan'' denotes the controller trained on deterministic fast-scan responses with rescued slow labels, while ``Sampled-fast'' denotes the controller trained with sampled-fast augmentation. 
The main IDPR configuration uses the Answer-only Fast policy and the Deterministic-scan controller.
}
\label{tab:variant_analysis}
\end{table*}

Next, we compare variants of the IDPR system. The sampled fast augmented controller was designed to reduce distribution mismatch between deterministic fast scan responses and sampled fast responses used during policy refinement. The GRPO-refined fast policy was motivated by the possibility that the fast policy should be optimized under the inhibition control mechanism used at deployment.

Table~\ref{tab:variant_analysis} shows that the simplest configuration: answer-only fast policy with the deterministic-scan inhibition controller achieves the highest held-out accuracy among the tested variants. Replacing the deterministic-scan inhibition controller with the sampled-fast augmented inhibition controller yields comparable accuracy with lower average token cost, but does not improve held-out accuracy. Similarly, the GRPO-refined fast policy does not consistently outperform the answer-only fast policy. These results suggest that, in our setting, the main gains come from inhibition control and validation-selected operation point rather than from sampled-fast augmentation or GRPO refinement.

\subsection{Effect of Threshold Calibration}
The inhibition controller outputs a continuous switch score, and the threshold determines how aggressively IDPR suppresses fast responses and invokes slow reasoning. We compare the validation-selected accuracy-first threshold with a fixed target-rate threshold. The fixed threshold is substantially more conservative on the held-out test set: it invokes slow reasoning on only 2.10\% of examples, improves accuracy from 47.90\% to 48.20\% (+0.30), and uses 143.64 average tokens. By contrast, the accuracy-first threshold invokes slow reasoning on 8.20\% of examples, improves accuracy to 48.92\% (+1.02), and uses 519.42 average tokens.

Interestingly, the fixed threshold has similar corrective precision to the accuracy-first threshold, 27.62\% versus 27.07\%, suggesting that it selects high-quality switches but with low recall. Thus, fixed calibration is more cost-conservative, whereas accuracy-first validation calibration selects a higher-recall operating point that allows more examples to benefit from deliberation. This confirms that threshold calibration is not a minor implementation detail: it determines the deployed accuracy-cost trade-off of IDPR.

This behavior also clarifies the role of the controller score. The fixed threshold does not fail because it selects poor examples: its corrective precision remains comparable to that of the accuracy-first threshold. Instead, it fails to select enough examples to realize a larger accuracy gain. In other words, the controller score provides a useful ranking over fast responses, but the threshold determines how much of this ranking is converted into actual deliberation. This distinction is important for deployment. A conservative threshold may be preferable when latency or token cost is the primary constraint, whereas an accuracy-first threshold is more appropriate when improving correctness on difficult examples is the main objective.

\section{Conclusion}
We introduced IDPR, a framework for inhibitory deliberation. Instead of deciding whether to use slow reasoning solely from the input question, IDPR first generates a concise fast answer and then determines whether a specific response should be released or suppressed. This design treats the fast answer as a prepotent candidate response and uses an inhibition controller to estimate whether invoking a deliberative reasoner is likely to provide sufficient corrective benefit.

Our experiments held-out mathematical reasoning tasks show that IDPR improves accuracy under a bounded slow-call budget. With a slow-call rate of 8.20\%, IDPR improves accuracy from 47.90\% to 48.92\%, while outperforming random and confidence-based same budget routing baselines. Source-wise analysis further shows that the gain is larger on the harder MoT subset, where the inhibition controller allocates more slow reasoning calls and achieves higher corrective precision. These results suggest that this inhibition control provides a more targeted mechanism for selective slow reasoning  than generic uncertainty heuristics.

At the same time, our results highlight the cost of slow reasoning. IDPR is not intended to make slow reasoning free; rather, it makes the accuracy-cost trade-off explicit by reporting slow-call rate, generated-token cost, and cost-aware utility. This aligns with the cognition motivation of the framework: slow reasoning is useful but costly, and should be recruited selectively when its expected benefit justifies the additional effort.

More broadly, IDPR provides a simple mechanism for combining fast intuitive answer with slow reasoning. Future work can instantiate the framework with stronger fast and slow policies, improve calibration across distributions, and explore richer inhibition controller architecture in broader agentic reasoning settings.

\section*{Limitations}
This work studies the inhibition control mechanism used for selective slow reasoning in a mathematical reasoning setting. Although IDPR improves accuracy under a bounded slow-call budget, slow reasoning remains costly: the accuracy-first operating point increases generated-token cost and may reduce cost-aware utility relative to always using the fast answer. The results also depend on the chosen fast--slow model pair, answer parser, and task distribution; stronger direct solvers may achieve higher raw accuracy, and the inhibition threshold may require recalibration when models, decoding strategies, or data distributions change. Finally, our cognitive control framing should be understood as a computational analogy rather than a biological claim: IDPR captures the idea of forming a fast candidate response and deciding whether to release or suppress it, but does not model the full complexity of human cognitive control.

\bibliography{custom}

@article{chen2023frugalgpt,
  title={FrugalGPT: How to Use Large Language Models While Reducing Cost and Improving Performance},
  author={Chen, Lingjiao and Zaharia, Matei and Zou, James},
  journal={arXiv preprint arXiv:2305.05176},
  year={2023}
}

@inproceedings{ding2024hybrid,
  title={Hybrid LLM: Cost-Efficient and Quality-Aware Query Routing},
  author={Ding, Dujian and Mallick, Ankur and Wang, Chi and Sim, Robert and Mukherjee, Subhabrata and Ruhle, Victor and Lakshmanan, Laks V. S. and Awadallah, Ahmed Hassan},
  booktitle={International Conference on Learning Representations},
  year={2024}
}

@inproceedings{ong2025routellm,
  title={RouteLLM: Learning to Route LLMs with Preference Data},
  author={Ong, Isaac and Almahairi, Amjad and Wu, Vincent and Chiang, Wei-Lin and Wu, Tianhao and Gonzalez, Joseph E. and Kadous, M. Waleed and Stoica, Ion},
  booktitle={International Conference on Learning Representations},
  year={2025}
}

@article{evans2013dualprocess,
  title={Dual-Process Theories of Higher Cognition: Advancing the Debate},
  author={Evans, Jonathan St. B. T. and Stanovich, Keith E.},
  journal={Perspectives on Psychological Science},
  volume={8},
  number={3},
  pages={223--241},
  year={2013}
}

@book{kahneman2011thinking,
  title={Thinking, Fast and Slow},
  author={Kahneman, Daniel},
  publisher={Farrar, Straus and Giroux},
  year={2011}
}

@article{miller2001integrative,
  title={An Integrative Theory of Prefrontal Cortex Function},
  author={Miller, Earl K. and Cohen, Jonathan D.},
  journal={Annual Review of Neuroscience},
  volume={24},
  pages={167--202},
  year={2001}
}

@article{shenhav2013expected,
  title={The Expected Value of Control: An Integrative Theory of Anterior Cingulate Cortex Function},
  author={Shenhav, Amitai and Botvinick, Matthew M. and Cohen, Jonathan D.},
  journal={Neuron},
  volume={79},
  number={2},
  pages={217--240},
  year={2013}
}

@article{deepseekai2025deepseekr1,

  title={DeepSeek-R1: Incentivizing Reasoning Capability in LLMs via Reinforcement Learning},

  author={{DeepSeek-AI}},

  journal={arXiv preprint arXiv:2501.12948},

  year={2025}

}

@inproceedings{snell2025scalingtesttime,

  title={Scaling LLM Test-Time Compute Optimally Can Be More Effective Than Scaling Model Parameters},

  author={Snell, Charlie and Lee, Jaehoon and Xu, Kelvin and Kumar, Aviral},

  booktitle={International Conference on Learning Representations},

  year={2025}

}

@inproceedings{brown2025largelanguagemonkeys,

  title={Large Language Monkeys: Scaling Inference Compute with Repeated Sampling},

  author={Brown, Bradley and Juravsky, Jordan and Ehrlich, Ryan and Clark, Ronald and Le, Quoc V. and R{\'e}, Christopher and Mirhoseini, Azalia},

  booktitle={International Conference on Learning Representations},

  year={2025}

}

@article{botvinick2001conflict,
  title={Conflict Monitoring and Cognitive Control},
  author={Botvinick, Matthew M. and Braver, Todd S. and Barch, Deanna M. and Carter, Cameron S. and Cohen, Jonathan D.},
  journal={Psychological Review},
  volume={108},
  number={3},
  pages={624--652},
  year={2001}
}

@article{aron2014inhibition,
  title={Inhibition and the Right Inferior Frontal Cortex: One Decade On},
  author={Aron, Adam R. and Robbins, Trevor W. and Poldrack, Russell A.},
  journal={Trends in Cognitive Sciences},
  volume={18},
  number={4},
  pages={177--185},
  year={2014}
}

@article{liu2025foundationagents,
  title={Advances and Challenges in Foundation Agents: From Brain-Inspired Intelligence to Evolutionary, Collaborative, and Safe Systems},
  author={Liu, Bang and Li, Xinfeng and Zhang, Jiayi and Wang, Jinlin and He, Tanjin and Hong, Sirui and Liu, Hongzhang and Zhang, Shaokun and Song, Kaitao and Zhu, Kunlun and others},
  journal={arXiv preprint arXiv:2504.01990},
  year={2025}
}

@article{kimi2025k15,

  title={Kimi k1.5: Scaling Reinforcement Learning with LLMs},

  author={{Kimi Team}},

  journal={arXiv preprint arXiv:2501.12599},

  year={2025}

}

@article{hu2025openreasonerzero,

  title={Open-Reasoner-Zero: An Open Source Approach to Scaling Up Reinforcement Learning on the Base Model},

  author={Hu, Jingcheng and Zhang, Yinmin and Han, Qi and Jiang, Daxin and Zhang, Xiangyu and Shum, Heung-Yeung},

  journal={arXiv preprint arXiv:2503.24290},

  year={2025}

}

@article{he2025skyworkor1,

  title={Skywork Open Reasoner 1 Technical Report},

  author={He, Jujie and Liu, Jiacai and Liu, Chris Yuhao and Yan, Rui and Wang, Chaojie and Peng, Cheng and Zhang, Xiaoyu and Zhang, Fuxiang and Xu, Jiacheng and Shen, Wei and others},

  journal={arXiv preprint arXiv:2505.22312},

  year={2025}

}

@article{ye2025limo,

  title={LIMO: Less is More for Reasoning},

  author={Ye, Yixin and Huang, Zhen and Xiao, Yang and Chern, Ethan and Xia, Shijie and Liu, Pengfei},

  journal={arXiv preprint arXiv:2502.03387},

  year={2025}

}

@article{abdin2025phi4reasoning,

  title={Phi-4-reasoning Technical Report},

  author={Abdin, Marah and Agarwal, Sahaj and Awadallah, Ahmed and Balachandran, Vidhisha and Behl, Harkirat and Chen, Lingjiao and de Rosa, Gustavo and Gunasekar, Suriya and Javaheripi, Mojan and Joshi, Neel and others},

  journal={arXiv preprint arXiv:2504.21318},

  year={2025}

}

@article{muennighoff2025s1,

  title={s1: Simple Test-Time Scaling},

  author={Muennighoff, Niklas and Yang, Zitong and Shi, Weijia and Li, Xiang Lisa and Fei-Fei, Li and Hajishirzi, Hannaneh and Zettlemoyer, Luke and Liang, Percy and Cand{\`e}s, Emmanuel and Hashimoto, Tatsunori},

  journal={arXiv preprint arXiv:2501.19393},

  year={2025}

}

@article{cobbe2021gsm8k,
  title={Training Verifiers to Solve Math Word Problems},
  author={Cobbe, Karl and Kosaraju, Vineet and Bavarian, Mohammad and Chen, Mark and Jun, Heewoo and Kaiser, Lukasz and Plappert, Matthias and Tworek, Jerry and Hilton, Jacob and Nakano, Reiichiro and Hesse, Christopher and Schulman, John},
  journal={arXiv preprint arXiv:2110.14168},
  year={2021}
}

@article{yang2024qwen25math,
  title={Qwen2.5-Math Technical Report: Toward Mathematical Expert Model via Self-Improvement},
  author={Yang, An and Zhang, Beichen and Hui, Binyuan and Gao, Bofei and Yu, Bowen and Li, Chengpeng and Liu, Dayiheng and Tu, Jianhong and Zhou, Jingren and Lin, Junyang and others},
  journal={arXiv preprint arXiv:2409.12122},
  year={2024}
}

@misc{openr1distill7b,
  title={OpenR1-Distill-7B},
  author={{Open R1 Team}},
  year={2025},
  howpublished={\url{https://huggingface.co/open-r1/OpenR1-Distill-7B}}
}

@misc{openr1,
    title = {Open R1: A fully open reproduction of DeepSeek-R1},
    url = {https://github.com/huggingface/open-r1},
    author = {Hugging Face},
    month = {January},
    year = {2025}
}

@article{kool2010decision,

  title={Decision Making and the Avoidance of Cognitive Demand},

  author={Kool, Wouter and McGuire, Joseph T. and Rosen, Zev B. and Botvinick, Matthew M.},

  journal={Journal of Experimental Psychology: General},

  volume={139},

  number={4},

  pages={665--682},

  year={2010}

}

@article{westbrook2013subjective,

  title={What Is the Subjective Cost of Cognitive Effort? Load, Trait, and Aging Effects Revealed by Economic Preference},

  author={Westbrook, Andrew and Kester, Daria and Braver, Todd S.},

  journal={PLOS ONE},

  volume={8},

  number={7},

  pages={e68210},

  year={2013}

}

@article{logan1984executive,
  title={On the Ability to Inhibit Thought and Action: A Theory of an Act of Control},
  author={Logan, Gordon D. and Cowan, William B.},
  journal={Psychological Review},
  volume={91},
  number={3},
  pages={295--327},
  year={1984},
  doi={10.1037/0033-295X.91.3.295}
}

\newpage
\appendix
\section{Appendix}
\label{sec:appendix}
\subsection{Switch Score Derivation}
\label{app:score_derivation}

In the main text, we define the switch score as
\[
S_\lambda
=
\widehat{\Delta Q}
-
\lambda_{\mathrm{inf}}\widehat C_s.
\]
Here we derive this score from the slow-over-fast utility difference.

If the fast answer is released, the utility is
\[
U_f
=
\alpha A_f
-
\lambda_{\mathrm{tok}}C_f
-
\rho F_f.
\]
If the slow branch is invoked after the fast response has already been generated, the utility becomes
\[
U_s
=
\alpha A_s
-
\lambda_{\mathrm{tok}}(C_f+C_s)
-
\rho F_s.
\]
Since \(C_f\) is shared by both decisions, it cancels out:
\[
U_s-U_f
=
\alpha(A_s-A_f)
-
\rho(F_s-F_f)
-
\lambda_{\mathrm{tok}}C_s.
\]
This motivates the quality-gain target
\[
\Delta Q
=
\alpha(A_s-A_f)
-
\rho(F_s-F_f),
\]
and the inference-time switch score
\[
S_\lambda
=
\widehat{\Delta Q}
-
\lambda_{\mathrm{inf}}\widehat C_s.
\]

\subsection{Fast-Side Evidence and Answer Parsing}
\label{app:fast_features}

Table~\ref{tab:fast_features} summarizes the fast-side evidence features used by the inhibition controller.

\begin{table*}[h]
\centering
\small
\begin{tabular}{lll}
\toprule
Category & Features & Description \\
\midrule
Confidence & Avg. log-probability, avg. logit margin & Token-level uncertainty \\
Parseability & Candidate count, answer form, format status & Whether the output is a valid final answer \\
Generation cost & Fast token count, latency, throughput & Cost and stability of fast generation \\
Degeneracy & Repetition score, truncation flag & Abnormal or incomplete generation \\
Source & Source indicator & Dataset/source group signal \\
\bottomrule
\end{tabular}
\caption{
Fast-side evidence features used by the inhibition controller.
}
\label{tab:fast_features}
\end{table*}

For a fast response \(y_f=(w_1,\ldots,w_T)\), the average token log-probability is
\[
\bar{\ell}_f
=
\frac{1}{T}
\sum_{t=1}^{T}
\log p_{\pi_f}(w_t \mid x,w_{<t}),
\]
and the average logit margin is
\[
m_f
=
\frac{1}{T}
\sum_{t=1}^{T}
(z_t^{(1)}-z_t^{(2)}),
\]
where \(z_t^{(1)}\) and \(z_t^{(2)}\) are the largest and second-largest logits at decoding step \(t\).

For answer extraction, we first search for an output line in the format \texttt{Final: <answer>}. If this is missing, we use fallback patterns such as boxed answers, GSM-style hash answers, or the last answer-like line. Outputs with no parseable answer, multiple conflicting answers, severe repetition, or truncation receive a format penalty.
\subsection{Router Pool Construction and Controller Training}
\label{app:router_pool}

The controller is trained from paired fast--slow outcomes. For each problem \(x\), we run both policies:
\[
y_f \sim \pi_f(\cdot\mid x),
\qquad
y_s \sim \pi_s(\cdot\mid x).
\]
The fast policy is decoded deterministically to produce a concise answer-only response. The slow policy produces a longer deliberative solution followed by a final answer. We evaluate both outputs against the gold answer and record correctness indicators \(A_f,A_s\), format penalties \(F_f,F_s\), and generated-token costs \(C_f,C_s\).

The controller is trained with a multi-task objective:
\[
\mathcal L_{\mathrm{ctrl}}
=
\mathcal L_Q
+
\beta_C\mathcal L_C
+
\beta_R\mathcal L_R,
\]
where \(\mathcal L_Q\) regresses the quality-gain target, \(\mathcal L_C\) regresses the slow-cost target, and \(\mathcal L_R\) is a binary classification loss for the corrective label
\[
y_{\mathrm{corr}}
=
\mathbbm{1}[A_f=0 \land A_s=1].
\]
Since corrective examples are sparse, we use positive weighting and weighted sampling for the corrective objective.

\end{document}